\nonstopmode
\pdfoutput=1
\documentclass{article}
\usepackage{spconf,amsmath,graphicx}
\usepackage{color,soul}
\usepackage{amsfonts}
\usepackage[bibstyle=ieee,backend=biber,firstinits=true]{biblatex}

\usepackage{bbm}
\usepackage{booktabs}
\usepackage[font={footnotesize}]{caption}
\usepackage{hyperref}
\usepackage{mathtools}
\usepackage{pgf}
\usepackage{rotating}
\usepackage{url}
\title{Very Deep Convolutional Networks for End-to-End Speech Recognition}
\name{Yu Zhang$^{1}\sthanks{Work done as Google Brain interns.}$, William Chan$^{2}$\footnotemark[1], Navdeep Jaitly$^{3}$}
\address{
  $^1$Massachusetts Institute of Technology \quad $^2$Carnegie Mellon University \quad $^3$Google Brain\\
  {\footnotesize \tt \texttt{yzhang87@mit.edu}, \texttt{williamchan@cmu.edu}, \texttt{ndjaitly@google.com}}
}
\begin{document}
\ninept
\maketitle

\begin{abstract}
Sequence-to-sequence models have shown success in end-to-end speech recognition.
However these models have only used shallow acoustic encoder networks. In our work,
we successively train very deep convolutional networks to add more expressive power and
better generalization for end-to-end ASR models. We apply network-in-network principles,
batch normalization, residual connections and convolutional LSTMs to build very deep
recurrent and convolutional structures. Our models exploit the spectral structure in the feature space
and add computational depth without overfitting issues.  We experiment with the WSJ
ASR task and achieve 10.5\% word error rate without any dictionary or language using a 15 layer deep network.
\end{abstract}

\section{Introduction}
\label{sec:introduction}
The sequence-to-sequence (seq2seq) model with attention \cite{bahdanau-iclr-2015}
has recently demonstrated a promising new direction for ASR that entirely sidesteps
the complicated machinery developed for classical ASR \cite{chorowski-nips-2015,
chan-icassp-2016,bahdanau-icassp-2016,bahdanau-iclr-2016,chan-interspeech-2016}. It is able to
do this because it is not restricted by the classical independence assumptions
of Hidden Markov Model (HMM) \cite{rabiner-ieeeproceedings-1989}
and Connectionist Temporal Classification (CTC) \cite{graves-icml-2012} models.
As a result, a single {\it{end-to-end model}} can jointly accomplish the ASR
task within one single large neural network.

The foundational work on seq2seq models, however, has relied on simple neural
network encoder and decoder models using recurrent models with
LSTMs \cite{bahdanau-icassp-2016,chan-interspeech-2016} or GRUs \cite{bahdanau-icassp-2016}. 
However, their use of hierarchy in the encoders demonstrates that better encoder
networks in the model should lead to better results. In this work we significantly extend
the state of the art in this area by developing very deep hybrid convolutional and
recurrent models, using recent developments in the vision community.

Convolutional Neural Networks (CNNs) \cite{lecun-ieeeproceedings-1998} have
been successfully applied to many ASR tasks
\cite{abdelhamid-interspeech-2013,sainath-icassp-2013,chan-icassp-2015}.
Unlike Deep Neural Networks (DNNs) \cite{hinton-ieeespm-2012}, CNNs explicitly exploit
structural locality in the spectral feature space. CNNs use shared weight
filters and pooling to give the model better spectral and temporal invariance
properties, thus typically yield better generalized and more robust models compared
to DNNs \cite{sainath-asru-2013}. Recently, very deep CNNs architectures
\cite{simonyan-iclr-2015} have also been shown to be successful in ASR
\cite{sercu-icassp-2016,sercu-interspeech-2016}, using more non-linearities,
but fewer parameters. Such a strategy can lead to more expressive models with
better generalization.

While very deep CNNs have been successfully applied to ASR, recently there have
been several advancements in the computer vision community on very deep CNNs
\cite{simonyan-iclr-2015,szegedy-cvpr-2015} that have not been explored in the
speech community. We explore and apply some of these techniques in our
end-to-end speech model:
\begin{enumerate}
  \item Network-in-Network (NiN) \cite{lin-iclr-2014} increases network depth
through the use of 1x1 convolutions. This allows us to increase the depth and
expressive power of a network while reducing the total number of parameters that
would have been needed otherwise to build such deeper models. NiN has seen great
success in computer vision, building very deep models\cite{szegedy-cvpr-2015}.
We show how to apply NiN principles in hierarchical Recurrent Neural Networks
(RNNs) \cite{hihi-nips-1996}.

  \item  Batch Normalization (BN) \cite{ioffe-icml-2015} normalizes each layer's
inputs to reduce internal covariate shift. BN speeds up training and acts as an
regularizer. BN has also seen success in end-to-end CTC models
\cite{amodei-icml-2016}. The seq2seq attention mechanism
\cite{bahdanau-iclr-2015} has high variance in the gradient (especially from
random initialization); without BN we were unable to train the deeper seq2seq
models we demonstrate in this paper. We extend on previous work and show how
BN can be applied to seq2seq acoustic model encoders.

  \item Residual Networks (ResNets) \cite{he-cvpr-2016} learns a residual
function of the input through the usage of skip connections. ResNets allow us
to train very deep networks without suffering from poor optimization or
generalization which typically happen when the network is trapped at a local
minima. We explore these skip connections to build deeper acoustic encoders.

  \item Convolutional LSTM (ConvLSTM) \cite{shi-nips-2015} use convolutions to replace the inner products within the LSTM unit.
ConvLSTM allows us to maintain structural representations in our cell state and output. Additionally, it allows us to add more compute to the model while reducing the number of parameters for better generalization. We show how ConvLSTMs can be beneficial and replace LSTMs.
\end{enumerate}
We are driven by same motivation that led to the success of very deep networks
in vision \cite{simonyan-iclr-2015,szegedy-cvpr-2015,ioffe-icml-2015,he-cvpr-2016} --
add depth of processing using more non-linearities and expressive power, while
keeping the number of parameters manageable, in effect increasing the amount of
computation per parameter. In this paper, we use very deep CNN techniques to
significantly improve over previous shallow seq2seq speech recognition models
\cite{bahdanau-icassp-2016}. Our best model achieves a WER of 10.53\% where our
baseline acheives a WER of 14.76\%.  We present detailed analysis on how each
technique improves the overall performance.

\section{Model}
In this section, we will describe the details of each component of our model.

\subsection{Listen, Attend and Spell}
Listen, Attend and Spell (LAS) \cite{chan-icassp-2016} is an attention-based seq2seq model which learns to
transcribe an audio sequence to a word sequence, one character at a time.
Let $\mathbf{x} = (x_1, \ldots, x_T)$ be the input sequence of audio frames, and $\mathbf{y} =
(y_1, \ldots, y_S)$ be the output sequence of characters. The LAS models each
character output $y_i$ using a conditional distribution over the previously emitted
characters $y_{<i}$ and the input signal $\mathrm{\bold x}$. The probability of the
entire output sequence is computed using the chain rule of probabilities:
\begin{eqnarray*}
  P(\mathbf{y} | \mathbf{x}) = \prod_i P({y}_i | \mathbf{x}, \mathbf{y_{<i}})
\end{eqnarray*}
The LAS model consists of two sub-modules: the listener and the speller. The
listener is an acoustic model encoder and the speller is an attention-based
character decoder. The encoder (the $\mathrm{Listen}$ function) transforms the original signal
$\mathbf{x}$ into a high level representation $\mathbf{h} = (h_1,\ldots, h_U)$
with $U \leq T$. The decoder (the $\mathrm{AttendAndSpell}$ function) consumes $\mathbf h$ and
produces a probability distribution over character sequences: 
\begin{eqnarray}
  \mathbf{h} &=& \mathrm{Listen}(\mathbf{x}) \\
  P(\mathbf{y}| \mathbf{x}) &=& \mathrm{AttendAndSpell}(\mathbf{h})
\end{eqnarray}
The $\mathrm{Listen}$ is a stacked Bidirectional Long-Short Term Memory (BLSTM)
\cite{graves-asru-2013} network with hierarchical subsampling as described in
\cite{chan-icassp-2016}. In our work, we replace $\mathrm{Listen}$ with a
network of very deep CNNs and BLSTMs.
The $\mathrm{AttendAndSpell}$ is an attention-based transducer
\cite{bahdanau-iclr-2015}, which generates one character $y_i$ at a time:
\begin{align}
	s_i &= \mathrm{DecodeRNN}([y_{i - 1}, c_{i - 1}], s_{i - 1}) \\
	c_i &= \mathrm{AttentionContext}(s_i, \mathbf h) \\
  p(y_i | \mathbf{x},  \mathbf{y_{<i}}) &= \mathrm{TokenDistribution}(s_i, c_i)
\end{align}
The $\mathrm{DecodeRNN}$ produces a transducer state $s_i$ as a function of the previously emitted token $y_{i-1}$, the previous attention context $c_{i - 1}$,
and the previous transducer state $s_{i - 1}$. In our implementation, $\mathrm{DecodeRNN}$ is a LSTM \cite{hochreiter-neuralcomputation-1997} function without peephole connections.

The $\mathrm{AttentionContext}$ function generates $c_i$ with a content-based Multi-Layer Perceptron (MLP) attention network \cite{bahdanau-iclr-2015}. 
\subsection{Network in Network}
\label{sec:NiN}
In our study, we add depth through NiN modules in the hierarchical subsampling connections between LSTM layers. 
We introduce a projected subsampling layer, wherein we simply concatenate two time frames to a single frame,
project into a lower dimension and apply BN and ReLU non-linearity to replace the skip subsampling connections in \cite{chan-icassp-2016}. Moreover, we further increase the depth of the network by adding more NiN 1 $\times$ 1 concolution modules inbetween each LSTM layer. 
\subsection{Convolutional Layers} 
Unlike fully connected layers, Convolutional Neural Networks (CNNs) take into
account the input topology, and are designed to reduce translational variance
by using weight sharing with convolutional filters. CNNs have shown improvement
over traditional fully-connected deep neural networks on many ASR tasks
\cite{sainath-asru-2013, chan-icassp-2015}, we investigate the
effect of convolutional layers in seq2seq models.

In a hybrid system,  convolutions require the addition of context window for each frame, or a way to treat the full utterance as a single sample \cite{sercu-interspeech-2016}. 
One advantage of the seq2seq model is that the encoder can compute gradients over an entire utterance at once.
Moreover, strided convolutions are an essential element of CNNs. For LAS applying striding is also a natural way to reduce temporal resolution. 




\subsection{Batch Normalization}
\label{sec:BN}
Batch normalization (BN) \cite{ioffe-icml-2015} is a technique to accelerate training and improve
generalization, which is widely used in the computer vision community. 
Given a layer with output $\mathbf{x}$, BN is implemented by normalizing each layer's inputs:
\begin{equation}
  \mathrm{BN}(\mathbf{x}) = \gamma \frac{\mathbf{x} - \mathrm{E} [\mathbf{x}]}{(\mathrm{Var}[\mathbf{x}] + \epsilon)^{\frac{1}{2}}} + \beta
  \label{eq:bn}
\end{equation}
where $\gamma$ and $\beta$ are learnable parameters. 
The standard formulation of BN for CNNs can be readily applied to DNN acoustic
models and cross-entropy training. 
For our seq2seq model, since we construct a minibatch containing multiple utterances, we follow the sequence-wise normalization \cite{amodei-icml-2016}. 
For each output channel, we compute the mean and variance statistics across all timesteps in the minibatch.

\subsection{Convolutional LSTM}
\begin{figure}[htb]
  \centering
  \includegraphics[width=0.6\linewidth]{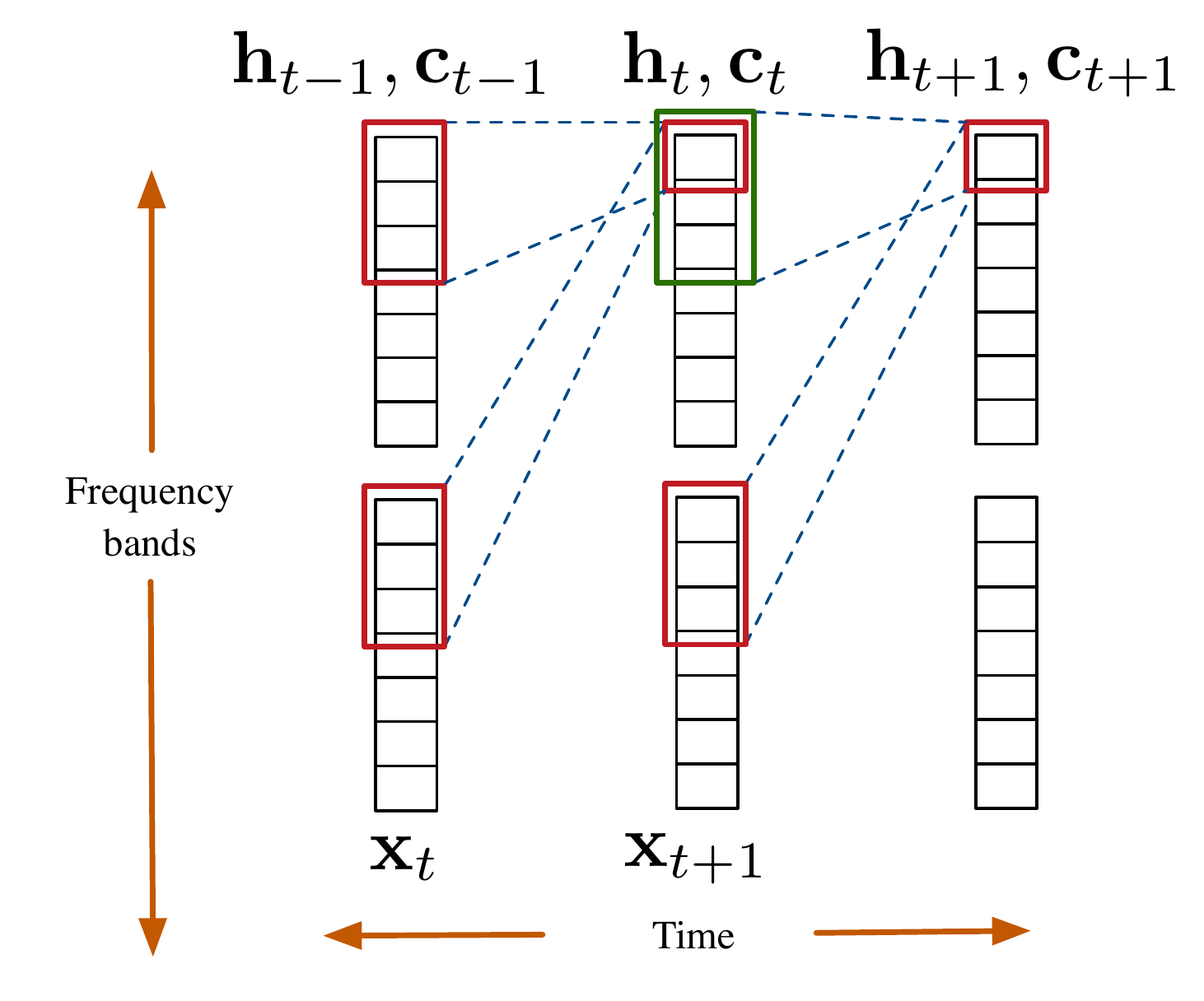}
  \caption{\footnotesize The Convolutional LSTM (ConvLSTM) maintains spectral structural localitly in its representation. We replace the inner product of the LSTM with convolutions.}
  \label{fig:convlstm}
\end{figure}

The Convolutional LSTM (ConvLSTM) was first introduced in \cite{shi-nips-2015}. Although the fully connected LSTM layer has proven powerful for handling temporal correlations, 
it cannot maintain structural locality, and is more prone to overfitting. ConvLSTM is an extension of FC-LSTM which has convolutional strucutres in both the
input-to-state and state-to-state transitions:
\begin{align}
    \mathbf{i}_t &= \sigma (\mathbf{W}_{xi}\ast \mathbf{x}_t + \mathbf{W}_{hi}\ast \mathbf{h}_{t-1} +\mathbf{b}_i)\nonumber\\
    \mathbf{f}_t &= \sigma (\mathbf{W}_{xf}\ast \mathbf{x}_t + \mathbf{W}_{hf}\ast \mathbf{h}_{t-1} +\mathbf{b}_f)\nonumber\\
    \mathbf{c}_t &= \mathbf{f}_t\odot \mathbf{c}_{t-1} + \mathbf{i}_t \odot \tanh (\mathbf{W}_{xc}\ast \mathbf{x}_t + \mathbf{W}_{hc}\ast \mathbf{h}_{t-1}+\mathbf{b}_c)\nonumber\\
    \mathbf{o}_t &= \sigma(\mathbf{W}_{xo}\ast \mathbf{x}_t + \mathbf{W}_{ho}\ast \mathbf{h}_{t-1} + \mathbf{b}_o)\nonumber\\
    \mathbf{h}_t &= \mathbf{o}_t\odot \tanh(\mathbf{c}_t) \label{eq:lstm5}
\end{align}
iteratively from $t=1$ to $t=T$, where $\sigma(\dot)$ is the logistic sigmoid function, 
$\mathbf{i}_t, \mathbf{f}_t,\mathbf{o}_t,\mathbf{c}_t$ and $\mathbf{h}_t$ are vectors to represent values of the input gate, forget gate, output gate, cell activation, 
and cell output at time $t$, respectively. $\odot$ denotes element-wise product of vectors. $\mathbf{W}_{*}$ are the filter matrices connecting different gates, 
and $\mathbf{b}_{*}$ are the corresponding bias vectors. The key difference is that $\ast$ is now a convolution, while in a regular LSTM $\ast$ is a matrix multiplication. 
Figure \ref{fig:convlstm} shows the internal structure of a convolutional LSTM. The state-to-state and input-to-state transitions can be achieved by a convolutional operation (here we ignore the multiple input/output channels). 
To ensure the attention mechanism can find the relation between encoder output and the test embedding, FC-LSTM is still necessary. 
However, we can use these ConvLSTMs to build deeper convolutional LSTM networks before the FC-LSTM layers. 
We expect this type of layer to learn better temporal representations compared to purely convolutional layers while being less prone to overfitting than FC-LSTM layers.
We found bidirectional convolutional LSTMs to consistently perform better than unidirectional layers. All experiments reported in
this paper used bidirectional models; here on we use convLSTM to mean bidirectional convLSTM.

\subsection{Residual Network}

\begin{figure}[htb]
\begin{minipage}[b]{.48\linewidth}
  \centering
  \centerline{\includegraphics[width=3.0cm]{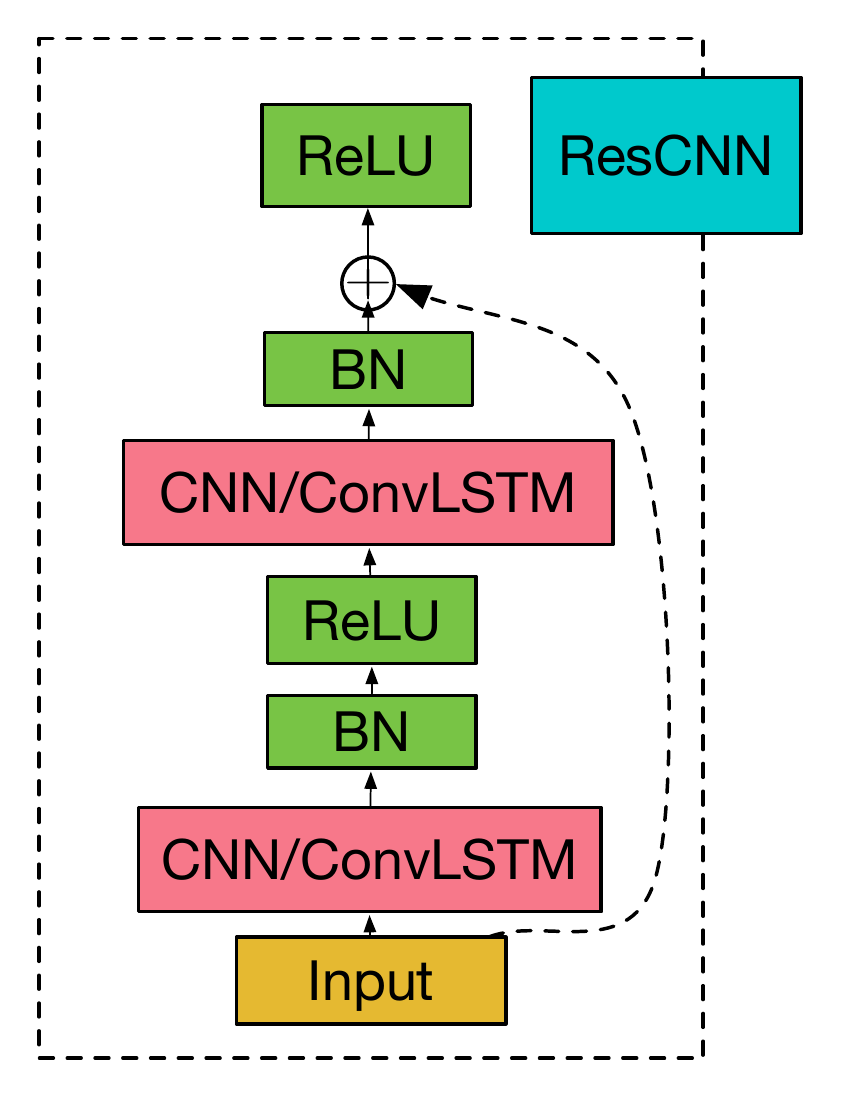}}
  \centerline{(a) ResCNN block}\medskip
\end{minipage}
\hfill
\begin{minipage}[b]{0.48\linewidth}
  \centering
  \centerline{\includegraphics[width=3.0cm]{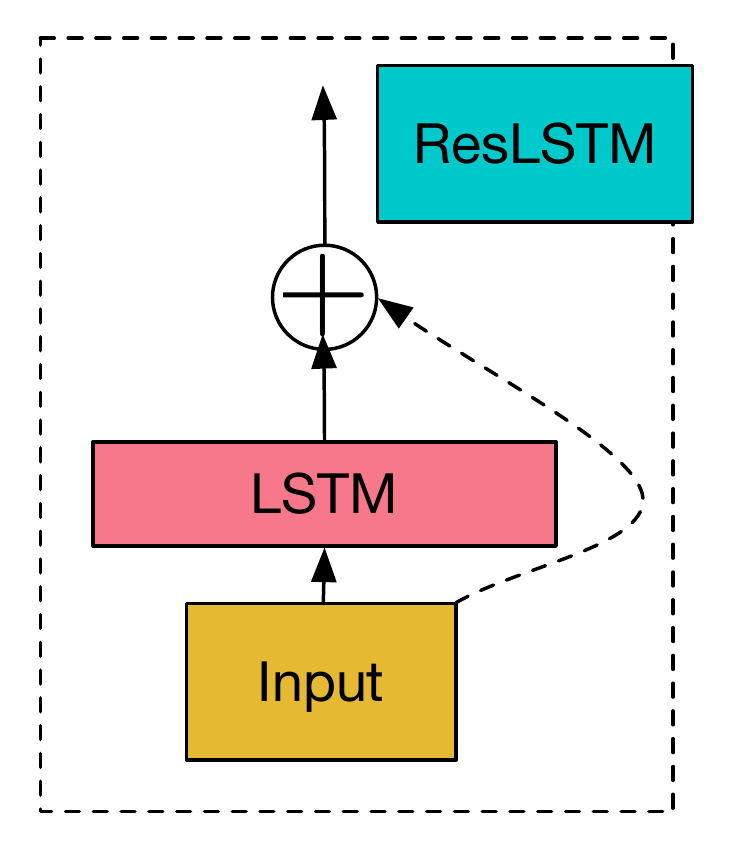}}
  \centerline{(b) ResLSTM block}\medskip
\end{minipage}
\caption{Residual block for different layers. ResCNN is a CNN block with CNN or ConvLSTM, Batch Normalization (BN) and ReLU non-linearities. The ResLSTM is a LSTM block with residual connections.}
\label{fig:resblock}
\end{figure}
Deeper networks usually improve generalization and often outperform shallow networks. 
However, they tend to be harder to train and slower to converge when the model becomes very deep. 
Several architectures have been proposed recently to enable training of very deep
networks~\cite{he-cvpr-2016, zhang-icassp-2016, kalchbrenner2015grid, srivastava2015training}.
The idea behind these approaches is similar to the LSTM innovation -- the introduction of linear or gated linear dependence between adjacent layers in the NN model to solve the vanishing gradient problem. 
In this study, we use a residual CNN/LSTM, to train deeper networks. 
Residual network \cite{he-cvpr-2016}  contains direct links between the lower layer outputs and the higher layer inputs. It defines a building block:
\begin{eqnarray}
\mathbf{y} = \mathcal{F} (\mathbf{x}, \mathbf{W}_i) + \mathbf{x}
\end{eqnarray}
where $\mathbf{x}$ and $\mathbf{y}$ are the input and output vectors of the layers considered. The function $\mathcal{F}$ can be one or more convolutional or convLSTM layers. 
The residual block for different layers is illustrated in Figure \ref{fig:resblock}. In our experiments, the convolutional based residual block always has a skip connection. However, for the LSTM layers we did not find skip connections necessary. 
All of the layers use the identity shortcut, and we did not find projection shortcuts to be helpful. 

\section{Experiments}
\label{sec:exp}
We experimented with the Wall Street Journal (WSJ) ASR task. We used the
standard configuration si284 dataset for training, dev93 for
validation and eval92 for test evaluation. Our input features were 80
dimensional filterbanks computed every 10ms with delta and delta-delta
acceleration normalized with per speaker mean and variance.
The baseline $\mathrm{EncodeRNN}$ function is a 3 layer
BLSTM with 256 LSTM units per-direction (or 512 total) and $4=2^2$ time factor
reduction. The $\mathrm{DecodeRNN}$ is a 1 layer LSTM with 256 LSTM units. All
the weight matrices were initialized with a uniform distribution
$\mathcal{U}(-0.1, 0.1)$ and bias vectors to $0$. For the convolutional model, 
all the filter matrices were initialized with a truncated normal distribution $\mathcal{N}(0, 0.1)$,
and used $32$ output channels. 
Gradient norm clipping
to $1$ was applied, together with Gaussian weight noise $\mathcal{N}(0, 0.075)$ and L2 weight
decay $1\mathrm{e}{-5}$ \cite{graves-nips-2011}. We used ADAM with the default
hyperparameters described in \cite{kingma-iclr-2015}, however we decayed the
learning rate from $1\mathrm{e}{-3}$ to $1\mathrm{e}{-4}$ after it converged. We used $10$ GPU
workers for asynchronous SGD under the TensorFlow framework
\cite{tensorflow2015-whitepaper}. We monitor the dev93 Word Error Rate (WER)
until convergence and report the corresponding eval92 WER.  The models took $O(5)$
days to converge.

\subsection{Acronyms for different type of layers}
All the residual block follow the structure of Fig. \ref{fig:resblock}.
Here are the acronyms for each component we use in the following subsections:
\begin{description}
  \item[P / 2] subsampling projection layer.  
  \item[C (f $\times$ t)] convolutional layer with filter f and t under frequency and time axis. 
  \item[B] batch normalization  
  \item[L] bidirectional LSTM layer. 
  \item[ResCNN] residual block with convolutional layer inside. 
  \item[ResConvLSTM] residual block with convolutional LSTM layer inside.
\end{description}

\subsection{Network in Network for Hierarchical Connections}
\label{subsec:nin}
We first begin by investigating the acoustic encoder depth of the baseline
model without using any convolutional layers. 
Our baseline
model follows \cite{bahdanau-icassp-2016} using the skip connection technique
in its time reduction. The baseline L $\times$ 3 or 3 layer BLSTM acoustic
encoder, model achieves a $14.76\%$ WER.

When we simply increase the acoustic model encoder depth (i.e., to depth 8),
the model does not converge well and we suspect the network to be trapped in
poor local minimas.  
By using the projection subsampling layer as discussed in Section \ref{sec:NiN}, we improves our WER to $13.61\%$ WER or a $7.8\%$ relative gain over the baseline.

We can further increase the depth of the network by adding more NiN
1 $\times$ 1 convolution modules inbetween each LSTM layer. This improves our
model's performance further to $12.88\%$ WER or $12.7\%$ relative over
the baseline. The BN layers were critical, and without them we found the model did
not converge well.  Table \ref{tab:NiN}
summarizes the results of applying network-in-network modules in the
hierarchical subsampling process. 

\begin{table}[htb]
  \centering
  \begin{tabular}{lc}
    \toprule
    Model & \bfseries  WER  \\
    \midrule
    L $\times$ 3 & 14.76  \\
    L $\times$ 8 & Diverged \\
    (L + P / 2 + B + R) $\times$ 2 + L & 13.61 \\
    (L + P / 2 + B + R + C(1$\times$1) + BN + R) $\times$ 2 + L & 12.88 \\
    \bottomrule
  \end{tabular}
  \caption{We build deeper encoder networks by adding NiN modules inbetween LSTM layers.}
  \label{tab:NiN}
\end{table}

\subsection{Going Deeper with Convolutions and Residual Connections}
In this subsection, we extend on Section \ref{subsec:nin} and describe
experiments in which we build deeper encoders by stacking convolutional layers and
residual blocks in the acoustic encoder before the BLSTM.  Unlike computer
vision applications or truncated BPTT training in ASR, seq2seq models need to
handle very long utterances (i.e., \textgreater2000 frames).  If we
simply stack a CNN before the BLSTMs, we quickly run out of GPU memory for deep
models and also have excessive computation times.  Our strategy to
alleviate this problem is to apply striding in the first and second layer of
the CNNs to reduce the time dimensionality and memory footprint.

We found no gains by simply stacking additional ResLSTM blocks even up to 8
layers. However, we do find gains if we use convolutions. If we stack 2
additional layers of 3 $\times$ 3 convolutions our model improves to 11.80\%
WER or 20\% relative over the baseline. If we take this model and add 8 residual
blocks (for a total of $(2 + (8) 2 + 5) = 23$ layers in the encoder) our model
further improves to $11.11\%$ WER, or a $24.7\%$ relative improvement over the baseline. We found that
using 8 residual blocks a slightly outperform 4 residual blocks.
Table \ref{tab:conv} summarizes the results of these experiments.
\begin{table}[htb]
  \centering
  \begin{tabular}{lc}
    \toprule
    Model & \bfseries  WER  \\
    \midrule
    L $\times$ 3 & 14.76  \\
    NiN (from Section \ref{subsec:nin}) & 12.88 \\
    \midrule
    ResLSTM $\times$ 8 & 15.00 \\
    (C (3 $\times$ 3) / 2) $\times$ 2 + NiN & 11.80 \\
    (C (3 $\times$ 3) / 2) $\times$ 2 + ResCNN $\times$ 4 + NiN & 11.30 \\
    (C (3 $\times$ 3) / 2) $\times$ 2 + ResCNN $\times$ 8 + NiN & 11.11 \\
    \bottomrule
  \end{tabular}
  \caption{We build deeper encoder networks by adding convolution and residual network blocks. The NiN block equals (L + C (1x1) + B + R) $\times$ 2 + L).}
  \label{tab:conv}
\end{table}

\subsection{Convolutional LSTM}
\begin{figure}[htb]
  \centering
  \includegraphics[width=0.6\linewidth]{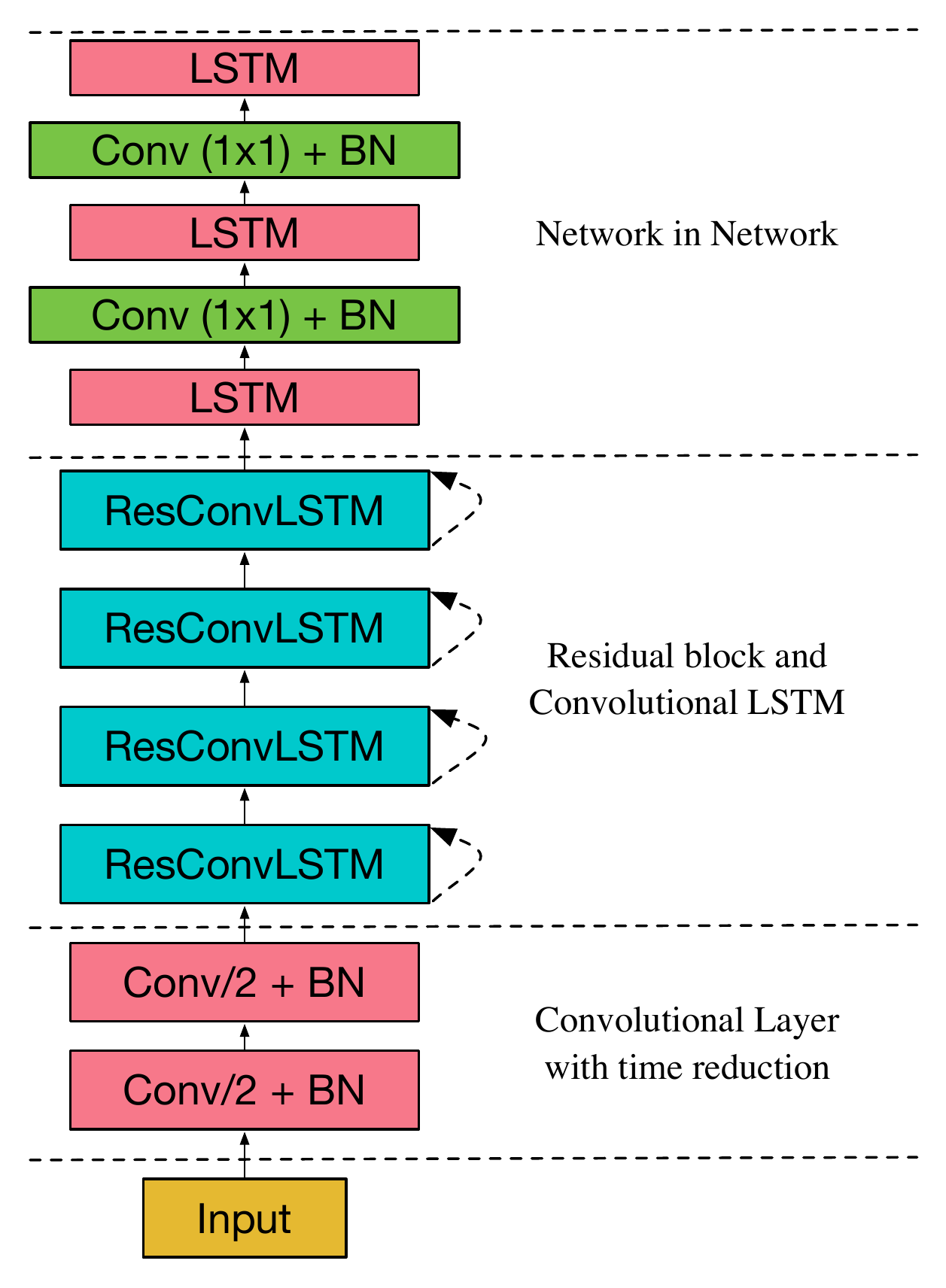}
  \caption{Our best model: includes two convolutional layer at the bottom and followed by four residual block and LSTM NiN block. Each residual block contains one convolutional LSTM layer and one convolutional layer. }
  \label{fig:bestmodel}
\end{figure}
\begin{table}[h]
  \centering
  \begin{tabular}{lc}
    \toprule
    Model & \bfseries  WER  \\
    \midrule
    L $\times$ 3 & 14.76  \\
    ConvLSTM $\times$ 3 & 24.23 \\
    \midrule
    (C (3$\times$3)) $\times$ 2 + ResCNN $\times$ 4 + NiN & 11.30 \\
    (C (3$\times$3)) $\times$ 2 + ResConvLSTM ($3\times1$) $\times$ 4 + NiN & 10.53 \\
    \bottomrule
  \end{tabular}
  \caption{Performance of models with convolutional LSTM layers. The NiN block equals (L + C (1x1) + B + R) $\times$ 2 + L).}
  \label{tab:convLSTM}
\end{table}
\begin{table}[htb]
  \centering
  \begin{tabular}{lc}
    \toprule
    \bfseries Model & \bfseries WER \\
    \midrule
    CTC (Graves et al., 2014) \cite{graves-icml-2014} & 30.1 \\
    seq2seq (Bahdanau et al., 2016) \cite{bahdanau-iclr-2016} & 18.0 \\
    seq2seq + deep convolutional (our work) & 10.53 \\
    \bottomrule
  \end{tabular}
  \caption{Wall Street Journal test eval92 Word Error Rate (WER) results across Connectionist Temporal Classification (CTC) and Sequence-to-sequence (seq2seq) models. The models were decoded without a dictionary or language model.}
  \label{tab:compare}
\end{table}

In this subsection, we investigate the effectiveness of the convolutional LSTM.
Table \ref{tab:convLSTM} compares the effect of using convolutional LSTM layers. It can be observed that a pure ConvLSTM performs much worse than the
baseline --- we still need the fully connected LSTM \footnote{We only use $32$ output channels thus it can be improve if we increase the channel size.}. However, replacing the
ResConv block with ResConvLSTM as shown in Figure \ref{fig:bestmodel} give us additional $7\%$ relative gains. 
In our experiments, we always use 3$\times$1 filters for ConvLSTM because the recurrent structure captures temporal information while the convolutions capture spectral structure. We conjecture that the gain is because the convolutional recurrent state maintains spectral structure and reduces overfitting.

Table \ref{tab:compare} compares our WSJ results with other published
end-to-end models. To our knowledge, the previous best reported WER on WSJ without an LM was the
seq2seq model with Task Loss Estimation achieving 18.0\% WER in \cite{bahdanau-iclr-2016}. Our baseline, also
a seq2seq model, achieved 14.76\% WER. Our model is different from that of~\cite{bahdanau-iclr-2016}
in that we did not use location-based priors on the attention model and we used weight
noise. Our best model, shown in Figure \ref{fig:bestmodel}, achieves a WER of 10.53\%.


\section{Conclusion}
\label{sec:conclusion}

We explored very deep CNNs for end-to-end speech recognition. We applied
Network-in-Network principles to add depth and non-linearities to hierarchical
RNNs. We also applied Batch Normalization and Residual connections to build
very deep convolutional towers to process the acoustic features. Finally, we
also explored Convolutional LSTMs, wherein we replaced the inner product of
LSTMs with convolutions to maintain spectral structure in its representation.
Together, we added more expressive capacity to build a very deep model without
substantially increasing the number of parameters. On the WSJ ASR task, we
obtained $10.5\%$ WER without a language model, an $8.5\%$ absolute improvement 
over published best result \cite{bahdanau-icassp-2016}. While we demonstrated our
results only on the seq2seq task, we believe this architecture should also
significantly help CTC and other recurrent acoustic models.

\vfill\pagebreak


\section{References}
\printbibliography[heading=none]
\end{document}